\documentclass[12pt]{article}

\usepackage{graphicx}
\usepackage{dsfont}
\usepackage{hyperref}
\usepackage{algorithm}
\usepackage{algpseudocode}
\usepackage{listings}
\usepackage{amsmath}
\usepackage{booktabs}
\usepackage{lipsum}
\usepackage{fullpage}
\usepackage{float}

\renewcommand{\u}{u}
\newcommand{\ut}{\frac{\partial \u}{\partial t}}
\newcommand{\ux}{\frac{\partial \u}{\partial x}}
\newcommand{\uxx}{\frac{\partial^2 \u}{\partial x^2}}

\newcommand{\loss}{\mathcal{L}}
\newcommand{\p}{\Theta}
\newcommand{\y}{y}
\newcommand{\barx}{\bar{x}}
\newcommand{\Barx}{\bar{X}}
\newcommand{\bary}{\bar{y}}
\newcommand{\Bary}{\bar{Y}}

\newcommand{\link}[1]{\href{https://#1}{#1}}

\usepackage{color}
\definecolor{deepblue}{rgb}{0,0,0.5}
\definecolor{deepred}{rgb}{0.6,0,0}
\definecolor{deepgreen}{rgb}{0,0.5,0}

\newcommand\pythonstyle{\lstset{
language=Python,
basicstyle=\scriptsize,
otherkeywords={self,True,False,Any,None,Callable,...},             
keywordstyle=\scriptsize\color{deepblue},
emph={},          
emphstyle=\scriptsize\color{deepred},    
commentstyle=\scriptsize\color{deepgreen},
frame=single,                         
showstringspaces=false            %
}}

\lstnewenvironment{python}[1][]
{
\pythonstyle
\lstset{#1}
}
{}

\title{PETScML: Second-order solvers for training regression problems in Scientific Machine Learning}
\author{Stefano Zampini$^1$, Umberto Zerbinati$^2$, George Turkiyyah$^1$, David Keyes$^1$}
\date{%
    $^1$Extreme Computing Research Center, King Abdullah University of Science and Technology\\%
    $^2$Mathematical Institute, University of Oxford\\[2ex]%
    \today
}
\begin{document}
\maketitle
\begin{abstract}
In recent years, we have witnessed the emergence of scientific machine learning as a data-driven tool for the analysis,
by means of deep-learning techniques, of data produced by computational science and engineering applications.

At the core of these methods is the supervised training algorithm to learn the neural network realization,
a highly non-convex optimization problem that is usually solved using stochastic gradient methods.
However, distinct from deep-learning practice, scientific machine-learning training problems feature a much larger volume
of smooth data and better characterizations of the empirical risk functions, which make them suited for conventional solvers
for unconstrained optimization.

We introduce a lightweight software framework built on top of the Portable and Extensible Toolkit for Scientific computation
to bridge the gap between deep-learning software and conventional solvers for unconstrained minimization.

We empirically demonstrate the superior efficacy of a trust region method based on the Gauss-Newton approximation
of the Hessian in improving the generalization errors arising from regression tasks when learning surrogate models
for a wide range of scientific machine-learning techniques and test cases. All the conventional second-order solvers tested,
including L-BFGS and inexact Newton with line-search, compare favorably, either in terms of cost or accuracy,
with the adaptive first-order methods used to validate the surrogate models.
\end{abstract}


\section{Introduction}

In recent years, there has been a growing interest in incorporating data-driven approaches into computational science and engineering, inspired by the advancements in deep-learning methods \cite{lecun2015deep,schmidhuber2015deep}.
This trend has given rise to the emerging discipline of scientific machine learning (SciML), which aims to tackle domain-specific data challenges by harnessing the predictive capabilities, interpretability, and domain knowledge offered by physics-based models.
One of the attractions of neural network models lies in their ability to handle high-dimensional function approximations \cite{cybenko1989, hornik1989multilayer};
this has led to the design of numerous techniques as scientific tools for uncovering the underlying physical laws hidden within experimental data.
Examples include the discovery of partial differential equations (PDEs) \cite{brunton2016discovering}, the learning of PDEs \cite{raissi2018deep, greenlearning_nature}, and PDE solvers \cite{raissi2019physics, yu2018deep}.
Noteworthily, deep-learning techniques have successfully pushed the boundaries of molecular dynamics simulations, enabling accurate simulations with hundreds of millions of atoms through \emph{ab initio} methods \cite{jia2020pushing},
and they have also been applied to solving the Hamilton--Jacobi--Bellman equations encountered in deterministic control problems \cite{nakamura2021adaptive}.

\subsection{Background}
Non-convex minimization problems are a fundamental challenge when training deep-learning models.
The loss landscape can be very complicated, given the
presence of multiple minima and saddle points, each possessing distinct generalization properties.
Second-order optimization methods applied to over-parametrized deep-learning training problems
have demonstrated overfitting tendencies, which hinder their generalization capabilities. This has led to
the widespread preference for stochastic first-order methods, as they naturally introduce regularization
in the stochastic regime, which helps to mitigate overfitting \cite{sgd_stoch_reg}.
Nevertheless, the landscape of deep
learning is ever-evolving, and in the era of vast datasets, we find ourselves in a highly informative data
regime for training these models. The sheer volume of data can potentially shift the
paradigm, creating a setting where overfitting might no longer severely impact generalization
capabilities, challenging conventional wisdom and opening doors to novel approaches.

Current practice for training deep-learning models is based on stochastic first-order methods that possess superior capabilities in finding local minima with better generalization properties for over-parametrized networks and data-scarce contexts,
for example gradient descent with momentum \cite{momentumSGD} and, to cite only a few, its adaptive relatives ADAM \cite{ADAM}, ADAMW \cite{ADAMW}, ADAGRAD \cite{ADAGRAD}.
Despite the appealing simplicity of these methods, their effectiveness depends on the training
task and it quickly reaches a plateau as larger parameter sizes are attempted.
Hyperparameter tuning, including setting the mini-batch size, the learning rate, and the annealing schedule, also depends on the learning task and lacks theoretical support.
In practice, grid-searching for a satisfactory set of hyper-parameters adds substantial expense when training new models.

On the other hand, given the large volume of data and better characterizations of the empirical risk functions, in the context of SciML
we may expect a smaller influence of the ``approximation'' and ``generalization'' errors on the approximation properties.
This, in turn, shifts the emphasis more towards the effective minimization of the ``training'' error, for which stochastic first-order methods are not the best choice.

\subsection{Related work}
Second-order methods have been extensively used and analyzed in the simulation and numerical
optimization literature and have succeeded in inverse problems in various
settings. Their dimension-independent convergence rate, robustness with respect to the condition
number of the Hessian matrix, and a much smaller set of hyper-parameters have made them the primary
tool for large-scale optimization in these areas. When it comes to deep-learning training, however, their
straightforward usage is relatively inefficient, and their memory and computational costs pose
fundamental challenges to their successful application.
This has led to several recent research works to
adapt these methods to the context and the extreme scales of current deep-learning.
Stochastic versions of quasi-Newton methods have been explored \cite{kaisa,goldfarb20,vandenbrand21,adahessian,nocedal18b,martensGN,rio18,shampoo,anil20,PSGD,KFAC,SENG,olearyroseberry21}
and have provided improved performance and robustness over first-order methods.
However, all these methods require modifications of the network implementation to store and operate on intermediate products.

The rise of deep-learning has paralleled impressive advances in the development of high-quality and highly efficient
open-source software frameworks. However, most of the focus has been on the design of the
various architectural network components and the efficient implementation of automatic differentiation for the
back-propagation algorithm. While first-order methods are relatively straightforward to implement,
second-order solvers have received little, if any,
attention from the deep-learning framework communities, and users of these frameworks are thus
left with a limited number of choices to train their models.
Much remains to follow early successes in these directions, above all in the context of matrix-free Newton-like methods,
where recent works have just analyzed their complexity and recently demonstrated their applicability to academic benchmark
problems \cite{newtonMR,TRs}.  For a recent effort to recast many of the stochastic Quasi-Newton methods cited above under the same computational framework, see \cite{ASDL}.

\subsection{Contributions}
Here, we are interested in matrix-free solvers because of their general applicability.
However, these solvers are characterized by a more complicated structure
requiring a deeper hierarchy of components, like line-search algorithms, Krylov methods, and preconditioners.
These components are at the core of high-performance libraries designed to solve large-scale
minimization problems; through carefully crafted application programming interfaces, the scientific community has extensively validated them on various applications, and they can offer a completely new arsenal
of methods to tackle the deep-learning training problem.
In this work, we present PETScML, which is a lightweight
Python interface exposing neural networks written using PyTorch \cite{pytorch} or JAX \cite{jax}
to the Portable and Extensible Toolkit for Scientific Computing (PETSc) \cite{petsc-web-page} and
its Python bindings module petsc4py \cite{petsc4py}, that allows quick experimentation with the many
different optimization solvers offered by PETSc.

To showcase the efficacy of PETScML, we focus on a particular source of
applications for regression tasks in SciML to empirically demonstrate
that second-order methods can improve trained model accuracies by superior exploitation of
the smoothness of the continuous fields and the richness of the datasets used, outperforming the hand-tuned
adaptive first-order methods used to validate the techniques.
In particular, we focus on the construction of inexpensive-to-evaluate surrogate models \cite{dino,FNO2,deeponet_natmachintell,greenlearning}
where networks are trained to learn the realization of the parameter-to-observable
maps used to solve inverse problems, Bayesian inversions, and
ultimately tackle the design-of-experiments framework. Improving the generalization accuracy of
surrogate neural operators can offer a large speed-up in these applications and enable a qualitatively new level
of training performance, accuracy, and robustness \cite{de2022cost}.

The paper is organized as follows. Section \ref{sec:training} introduces the minimization problem in training
deep-learning models and describes the matrix-free solvers studied in the paper. Section \ref{sec:petscml} gives
an overview of the software framework, while Section \ref{sec:results} contains numerical results testing
the solvers on a series of test cases taken from the recent literature.
Section \ref{sec:conclusions} provides further discussions and concludes the paper.

\section{Deep-learning training}\label{sec:training}

We are interested in the solution of the supervised learning problem, posed as the minimization of a non-convex scalar function
\begin{equation}\label{eq:training}
\arg \min_\p f(\p), \quad f(\p) = \frac{1}{N} \sum_{i=1}^{N} \loss ( \y(\p, \barx_i), \bary_i)
\end{equation}
where $N$ is the total number of data points, $\barx_i \in \Barx, \bary_i \in \Bary$ are the training points and expected outputs, $\y(\cdot, \cdot)$ the network prediction,
$\loss(\cdot,\cdot)$ a scalar convex ``loss'' function, and $\p$ the network parameters.

Usually, this minimization problem is initially recast to the so-called ``mini-batch'' framework by subsampling the data in $(\Barx, \Bary)$ to obtain a stochastic estimator of $f$.
Training is then performed using first-order methods, i.e.
\begin{equation*}
\p_{k+1} = \p_k - \lambda_k g_k, \quad g_k = \frac{1}{b} \sum_{i_k} \nabla_{\p_k} \loss ( \y(\p_k, \barx_{i_k}), \bary_{i_k}),
\end{equation*}
where $\lambda_k$ is the learning rate and $i_k$ a subset of the dataset of size $b$, i.e., the mini-batch size.
The above update is performed for a certain number of passes on the entire dataset, where each single pass is called ``epoch'' (i.e. an epoch is worth $N/b$ update steps).
Depending on the different variants, the stochastic gradient $g_k$ is scaled and adapted using momentum to accelerate convergence.

Since larger mini-batch sizes improve the arithmetic
intensity of the computations,
in an ideal world the only limiting factor for the selection of the mini-batch size should be
the amount of memory available on the computing devices, a very precious commodity.
However, the choice of the mini-batch size influences the convergence properties and the generalization capabilities of the stochastic methods.
Optimal configurations, from the point of view of the model accuracies, can lead to suboptimal utilization of the computing resources \cite{goyal2017accurate}.
For stochastic optimizers able to deal with large batch sizes, see \cite{lars,lamb}.

Here, we want to investigate if optimization methods of the type
\begin{equation*}
\p_{k+1} = \p_k - \lambda_k s_k, \quad s_k = \widetilde{H}_k^{-1}(\p_k) g_k,
\end{equation*}
can improve the generalization error and reduce the computational costs needed to converge to a local minimum.
$\widetilde{H}_k$ is some approximation of the symmetric indefinite Hessian $\frac{\partial^2 f}{\partial \p^2}$ and $\lambda_k$ may be computed by a line-search algorithm.
In particular, in this work, we consider three of the most widely used solvers for unconstrained minimization: the Limited memory Broyden--Fletcher--Goldfarb--Shanno Quasi-Newton method (L-BFGS, \cite{lbfgs}), inexact Newton with line-search (LS), and the trust region (TR) method.
Although stochastic variants of these methods have been proposed in the literature \cite{BFGSs,LSs,TRs}, here we only consider their deterministic, classical variants and solve the minimization problem given in \eqref{eq:training} directly.
A detailed description of these methods is outside the scope of this manuscript, and in what follows, we only discuss specific aspects that are relevant to this work; the interested reader can consult the excellent monographs \cite{NocedalAndWright,ConnGould} for additional details.

The L-BFGS method iteratively constructs a positive definite approximation of the inverse of the Hessian by storing and combining past iterates $(\p_k,g_k)$ using rank-2 updates.
The solution of the approximated problem is then combined with a line-search algorithm to guarantee descent in the objective function.
The major computational costs of the L-BFGS algorithm thus consist of evaluating the objective function during the line-search procedure and the gradient to update the L-BFGS approximation. In contrast, the cost of the linear algebra involved with applying the L-BFGS matrix inverse is usually negligible for large models and modest history sizes when using dense compact formulas \cite{denseQN}.

The LS and TR algorithms approximatively solve the linearized equations $\widetilde{H}_k^{-1} g_k$ by means of a Krylov method;
the cost of computing the action of the Hessian matrix on the vector becomes dominant, and the memory requirements increase
to store the additional computational graphs needed to evaluate higher-order derivatives.
Deciding to which accuracy to solve the update step is critical for the methods' efficiency, since they possess at most linear convergence rate when $\p_k$ is far from the solution, and the way they compute the update $s_k$ is substantially different.
LS solves the linear systems up to a dynamic relative tolerance
\begin{equation}\label{eq:ls_model}
\|g_k + \widetilde{H}_k s_k\| \leq \nu_k \|g_k\|,
\end{equation}
where $\nu_k$ is usually computed by looking at successive values of the norm of the gradient \cite{EisenstatWalker},
and then it combines the updated solution with a line-search algorithm, backtracking until a suitable descent in the objective function is found.
On the other hand, TR computes the direction of descent by solving the constrained quadratic model problem
\begin{equation}\label{eq:tr_model}
\arg \min_{s_k} m_k(s_k) := g^T_k s_k + \frac{1}{2} s^T_k \widetilde{H}_k s_k , \quad \mbox{s.t. } \|s_k\| \leq \Delta_k,
\end{equation}
where the size of the trust region $\Delta_k$ is dynamically adjusted according to the ratio between the reduction in the objective function and the model approximation, i.e.
\begin{equation}\label{eq:tr_accept}
\rho_k := \frac{f(\p_k) - f(\p_k + s_k)}{m_k(0) - m_k(s_k)}.
\end{equation}

Usually, the symmetric indefinite linear systems of equations are solved with a (preconditioned) conjugate gradient (CG) algorithm, which is halted whenever a negative eigenvalue (i.e., an ascent direction) is encountered.
Such stopping criterion can be evaluated based on the recurrence formulas of CG, and it can be considered as an extra criterion to prevent over-solving.
A simple modification of the CG algorithm, known as the Steihaug-Toint variant \cite{Steihaug,Toint}, can solve Eq. \eqref{eq:tr_model} approximatively.

A positive semi-definite approximation of the Hessian is also possible \cite{schraudolph,martensGN}, since Eq. \eqref{eq:training} admits a generalized Gauss-Newton (GN) approximation, defined as (dropping the $k$ subscript)
\begin{equation}\label{eq:gn}
\widetilde{H}_{GN} := J^T H_{\loss} J \,\quad J := \frac{\partial \y}{\partial \p} \,\quad H_{\loss} := \frac{1}{N}\frac{\partial^2 \loss}{\partial \y^2},
\end{equation}
which is routinely used to solve nonlinear least-squares problems. Such approximation results in a positive semi-definite linear operator provided the loss function is convex. In addition, the memory requirements of the Gauss-Newton approximation are similar in magnitude to those of the gradients.

\section{Software architecture} \label{sec:petscml}

PETScML \footnote{Repository will be made public before publication} is a lightweight Python interface built on top of petsc4py \cite{petsc4py},
the Python wrapper for PETSc \cite{petsc}, which is an award-winning software library widely used by the
computational science and engineering community to solve large-scale, time-dependent, nonlinear equations and optimization problems (via the Toolkit for Advanced Optimization, TAO).
PETSc is a performance portable framework that uses the Message Passing Interface (MPI) for distributed memory computations \cite{petsc-sf} and interfaces with different computational backends, including NVIDIA, AMD, and Intel GPUs \cite{petsc-gpu}.
All the solver configurations available are command-line customizable and require a small amount of custom user code.

At the highest level, PETScML defines an abstract class that exposes the basic methods needed by the optimization solvers:
\begin{python}
from PETSc import Vec, Mat
from MPI import Comm

class function:

    def __init__(self, params : Any, comm : Comm, device : str):
        ''' Perform the needed initialization.'''
        ...

    def objective(self, params : Any) -> float:
        ''' Return the objective value evaluated at params'''
        ...

    def gradient(self, params : Any) -> Vec:
        ''' Return the gradient evaluated at params'''
        ...

    def hessian(self, params : Any, gn : bool = False) -> Mat:
        ''' Return the Hessian operator evaluated at params,
            possibly using the Gauss-Newton approximation.
            Uses hessianMult or hessianMultGN'''
        ...

    def hessianMult(self, params: Any, input_vec : Any) -> Any:
        ''' Return H(params) @ input_vec. '''
        ...

    def hessianMultGN(self, params: Any, input_vec : Any) -> Any:
        ''' Return GN approximation H(params) @ input_vec. '''
        ...
\end{python}

The derived function classes implement these methods for each specific deep-learning backend,
which does not need to be aware of PETSc-specific details.
Parameter values can be in any form accepted by the backends: PETScML internally uses routines to flatten
parameter values into contiguous storage for vectors and, vice-versa, to fold vector data into the parameters
format needed by the backend.
The derived class must only implement the objective, gradient, and Hessian matrix-vector multiplication using their input
data format for the linearization point and the input vector.
Such an approach allows the separation of the public interface from the internal implementation.

Distributed memory is supported by specifying the relevant MPI communicator when initializing the function instance; device configurations supported are {\tt cpu} and {\tt gpu}.
NCCL, the NVIDIA collective communication library\footnote{Available at \link{developer.nvidia.com/nccl}} is
internally used for multi-GPU communications, while the exchange of information via
contiguous storage memory is handled using the no-copy protocol from dlpack\footnote{See \link{dmlc.github.io/dlpack/latest/}}.
The currently supported backends are JAX and PyTorch. Extensions to additional computational backends, like AMD or Intel GPUs, or Python ML backends
can be easily accomodated.

Models of the type given in \eqref{eq:training} are derived from the {\tt function} class:
\begin{python}
class mlModel(function):

    def __init__(self,
                 params: Any,             # parameters
                 y : Callable,            # model: y(params, x)
                 loss : Callable,         # loss(y(params,x), y)
                 train_loader : Iterable, # training dataset
                 **kwargs):
        ...

    def update(self, step : int) -> None:
        '''Perform the needed actions on the model
           and the dataset at the beginning of the step'''
        ...

\end{python}
These classes automatically compute objective function, gradient, and Hessian matrix-vector products using the user-provided model {\tt y}, {\tt loss} function, and training dataset {\tt train\_loader}.
The {\tt update} callback is used to load the subsequent batches and perform the needed model updates at the beginning of a new optimization step.
The loader can be any iterable Python type, such as a PyTorch {\tt DataLoader}; distributed loader classes are also provided to handle multi-device, data-parallel training.
Supported neural network module libraries for the JAX backend include DM-Haiku\footnote{Available at \link{dm-haiku.readthedocs.io/en/latest/}} and Flax\footnote{Available at \link{flax.readthedocs.io/en/latest/}}.

Optimization solvers can be created using the one-liner
\begin{python}
def createSolver(f : function,
                 cfg : dict = {},
                 params : Any = None,
                 monitor : Callable[[Any, int, float], None] = None,
                 prefix : str = None)
\end{python}
which takes as input one instance of the function to be minimized and optional keyword arguments for the
solver configuration dictionary {\tt cfg},
the initial values for the parameters {\tt params}, the convergence monitoring routine {\tt monitor},
and a string identifying the solver for the command-line customization {\tt prefix}.
For experimentation purposes, first-order stochastic methods from OPTAX\footnote{Available at \link{optax.readthedocs.io/en/latest/}} are wrapped as custom PETSc solvers;
they can be programmatically customized using the configuration dictionary {\tt cfg} or by command-line options.

\begin{table}
  \caption{Software}
  \label{tab:software}
  \begin{tabular}{ccl}
    \toprule
    Library & Version & Link\\
    \midrule
    PETSc & 3.21 & \link{gitlab.com/petsc/petsc}\\
    CUDA & 11.7 & \link{developer.nvidia.com/cuda-toolkit}\\
    JAX & 0.4.8 & \link{github.com/google/jax}\\
    PyTorch & 2.1.0 & \link{github.com/pytorch/pytorch}\\
    DeepXDE & 1.9.3 & \link{github.com/lululxvi/deepxde}\\
    OPTAX & 0.1.5 & \link{github.com/google-deepmind/optax}\\
    DM-Haiku & 0.0.9 & \link{github.com/google-deepmind/dm-haiku}\\
  \bottomrule
\end{tabular}
\end{table}
\section{Numerical results} \label{sec:results}

In this Section, we evaluate the performances of the L-BFGS, trust region, and inexact Newton with line-search solvers
on a series of test cases taken from the recent literature on SciML techniques for constructing surrogates to be used in inverse problems governed by PDEs;
the solvers' performances are always compared against the same adaptive first-order solvers used to validate the results, exactly matching the configurations reported in the literature.
To keep the focus on solvers' performances, we describe each of the SciML techniques in their own subsection; specific first-order solvers are always referred to as ``reference'' solvers,
and their performances are reported with a ``REF'' label in the figures.
For trust region and inexact Newton, we either consider an exact Hessian or its Gauss-Newton approximation (GN), as given in \eqref{eq:gn};
we remark that these matrices are never computed explicitly, but we only evaluate their action on given input vectors.

Convergence plots are reported for loss function values (i.e. the training error) and case-specific metrics (i.e. the generalization error) on the testing datasets; the minimum value attained throughout the minimization process is always shown in the legends.
As a qualitative comparison, the data is reported as a function of the epoch number to describe the convergence histories of the various solvers.
We also report the same data against the number of ``oracle calls'' performed by each solver as an architecture-agnostic measure of their costs, based on
the number of passes on the computational graph needed to perform the computations.
In particular, an objective function evaluation costs $1$ oracle call (one forward pass), a gradient costs $2$ oracle calls (one forward and one backward pass),
while an exact Hessian matrix-vector product costs $4$ calls since we need to differentiate the gradient operation.
In contrast, the Hessian matrix-vector product of the Gauss-Newton approximation \eqref{eq:gn} requires $2$ oracle calls; one forward pass to compute the action of $J$ and one backward pass to compute the action of $J^T$, while the computations associated with the Hessian of the loss function are negligible.
These costs are always rescaled, accounting for the relative size of the mini-batch with respect to the whole dataset.

We stress that these are only rough, memory-bandwith based estimates of the relative costs of the gradient and the Hessian matrix-vector products;
the actual computational costs depend on many factors including the computing architecture,
the arithmetic intensity of these operations, the implementation of automatic differentiation, the mini-batch size used,
the input pipeline's stochasticity, and the computational graph's complexity.
Given the high-degree of heterogeneity of the various testcases considered (network architectures, dataset sizes, and ML frameworks),
in what follows we will not discuss computational timings, code performances or scalability aspects.
For the interested reader, we can say that training times are always strongly correlated with the oracle calls measurements in all the cases studied.

All solvers are initialized with the same set of random model parameters, separately for each test case.
Convergence histories are reported on a log-log scale to facilitate cross-comparisons among solvers and test cases;
to this end, we also keep fixed the hyper-parameters of the PETScML solvers for all the numerical experiments.
In particular, we always use a simple, Armijo-type backtracking line-search for the L-BFGS, LS, and LS-GN solvers.
The number of history vectors stored for the L-BFGS approximation is $30$.
The dynamic computation of the relative tolerance $\nu_k$ for LS and LS-GN in Eq. \eqref{eq:ls_model} is computed using the procedure outlined in \cite{EW4} using a initial tolerance of $\nu_0 = 0.9$.
We also experimented with more classical variants \cite{EisenstatWalker}, but we found the one proposed in \cite{EW4} to be more robust.
We use standard parameters for the trust region methods; we accept a step when $\rho_k$ given in \eqref{eq:tr_accept} is larger than $0.001$. The trust region's size, initially set to $\delta_0 = 0.2$, is shrunk by a factor $0.25$ if $\rho_k < 0.25$ while it is doubled if $\rho_k > 0.75$. The maximum allowed trust region size is 10.
All linear systems are solved using the CG method for a maximum of $100$ iterations, using an inexpensive preconditioner constructed with an L-BFGS update with $5$ history vectors \cite{BFGSpre}.
The minimization procedure is halted whenever a line search fails, the size of the trust region can no longer be reduced further, or the prescribed maximum number of iterations is reached.

The relevant software libraries used for the numerical results are reported in Table \ref{tab:software}.
The script to reproduce the experiments, figures, and the driver applications for each specific test are public.
In all the test cases reproduced herein, we follow exactly the experimental settings used in the literature;
unless otherwise stated, the computations are performed in double precision floating point arithmetic and full-batch size configuration.
Numerical results have been obtained using A100 NVIDIA GPUs with 80 GB of device memory. The first-order solvers are always run using the OPTAX/JAX framework.

\subsection{Fourier Neural Operator} \label{sec:results_fno}

The Fourier Neural Operator (FNO) framework aims to learn mappings between infinite dimensional spaces using a finite collection of input/output pairs observations. More specifically, given two Banach spaces $\mathcal{A}, \mathcal{U}$ and a map $\mathcal{G}: \mathcal{A} \to \mathcal{U}$, the goal is to build an approximation $\mathcal{G}_\p$ using observations drawn from some probability measure supported on $\mathcal{A}$ using an empirical cost (or loss) function. Here $\mathcal{G}_\p$ is a neural network approximation of the form
\begin{equation*}
\mathcal{G}_\p := \mathcal{Q} \circ \sigma_L (\mathcal{K}_L) \circ \dots \circ \sigma_1 (\mathcal{K}_1) \circ \mathcal{P}
\end{equation*}
where $\mathcal{Q}$ and $\mathcal{P}$ are the ``projection'' and ``lifting'' operators, $\sigma_T$ an activation function acting component-wise on the output of some kernel operator $\mathcal{K}_T$.
For additional information, more rigorous definitions of functional settings, and different choices of the kernel operator, see \cite{FNO2}. For error analysis, see \cite{FNOuniversal}.

\subsubsection{Burgers' equations}  \label{sec:results_fno_burgers}
As a first test case, we consider the one-dimensional Burgers' equation test proposed in Section 5.1 of \cite{FNO}, where we want to learn the operator mapping the initial conditions to the final conditions at time $T=1$ of
\begin{equation*}
\ut + u~\ux = \nu \uxx, \quad x \in (0,1)
\end{equation*}
with $\nu=0.1$ the viscosity and with periodic boundary conditions. We use the same datasets as in \cite{FNO}, where the initial conditions are generated as Gaussian random fields while the final conditions are computed by an implicit method on a fine grid of size $8192$. Following \cite{deeponet_fair}, the data generated is then subsampled on a much coarser grid of size $128$; the training dataset consists of $1000$ pairs of random initial conditions and the associated final solution, while the testing dataset contains $100$ of such pairs. For the numerical results, we used the data provided by the authors\footnote{Available at \link{github.com/neuraloperator/neuraloperator/tree/master}}.
The network has an initial dense layer of $64$ features, followed by 4 Fourier layers using $16$ modes with a width equal to $64$, and a final dense layer with $128$ features, for a total of $549,569$ parameters. The activation function is the Gaussian Error Linear Units (GELU) function \cite{GELU}.
The network is implemented using DM-Haiku and the JAX backend; all parameters are stored as real numbers, and complex arithmetic in the Fourier layers is emulated in software.

\begin{figure}[H]
\centering
\includegraphics[width=\linewidth]{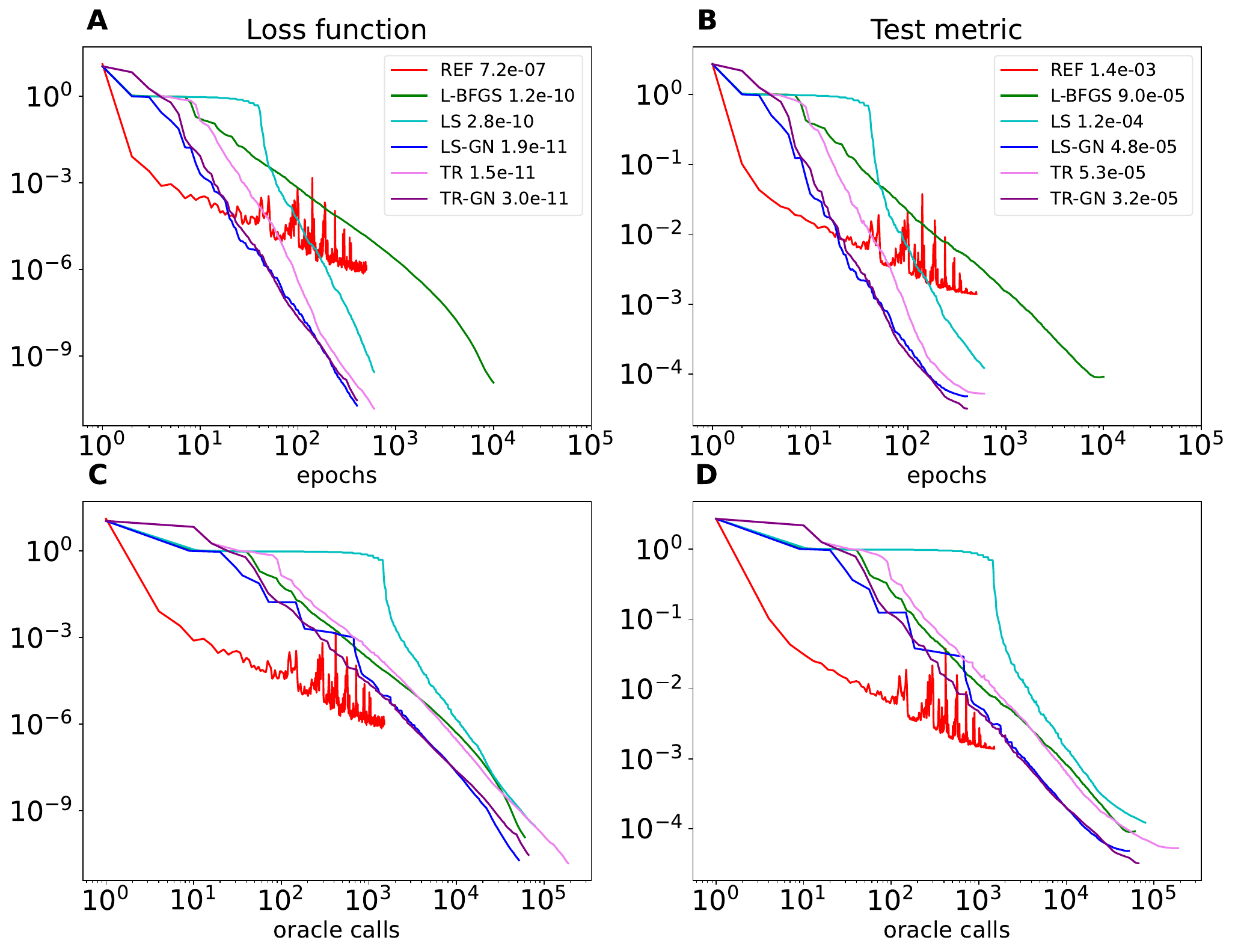}
\caption{FNO Burgers' test case. Convergence histories for loss function values and testing metrics in terms of epochs (top row, panels A and B) and oracle calls (bottom row, panels C and D) for different solvers. The values in the legends denote the minimum metric value achieved. See Section \ref{sec:results_fno_burgers} for details.}
\label{fig:fno_conv}
\end{figure}

The loss function is the mean squared $l2$ relative error
\begin{equation}\label{ref:msrelerr}
\loss(\y,\bary) = \frac{1}{N}\sum^N_{i=1}\frac{\|\y_i - \bary_i\|^2}{\|\bary_i\|^2},
\end{equation}
while the accuracy metric is the mean relative $l2$ error
\begin{equation}\label{ref:mrelerr}
\frac{1}{N}\sum^N_{i=1}\frac{\|\y_i - \bary_i\|}{\|\bary_i\|}.
\end{equation}
The solver used in \cite{FNO} is ADAMW with weight decay $0.0001$, mini-batch size 20, and initial learning rate $0.001$, halved every 100 epochs using an exponential staircase schedule.
Numerical results are reported in Fig. \ref{fig:fno_conv}.

The solvers tested exhibit different convergence histories regarding loss function values; the testing metric is always strongly correlated to the loss value, confirming that no overfitting is happening.
The reported test metric value for the reference solver is $0.0014$, one order of magnitude smaller than the one reported in \cite{deeponet_fair} but in line with \cite{FNO2} (see Table 3 in the reference).
The reference solver quickly reduces the loss value and the testing metric before entering a second phase of slow and erratic convergence to the identified local minimum, a typical behavior of stochastic first-order solvers.
On the other hand, all the other solvers exhibit an initial plateau phase before converging to a local minimum at different convergence rates regarding the number of epochs, with L-BFGS and the exact Hessian variants of TR and LS experiencing longer plateaus.
Despite their different convergence histories, they seem to converge to local minima with similar generalization errors, always at least one order of magnitude smaller than the reference solver, with TR-GN improving the generalization error of the reference solver by almost two orders of magnitude.
Analyzing the converge histories from the point of view of oracle calls, all the solvers have a similar cost, with the ``cheapest'' solver being TR-GN.

\begin{figure*}[h]
\centering
\includegraphics[width=\linewidth]{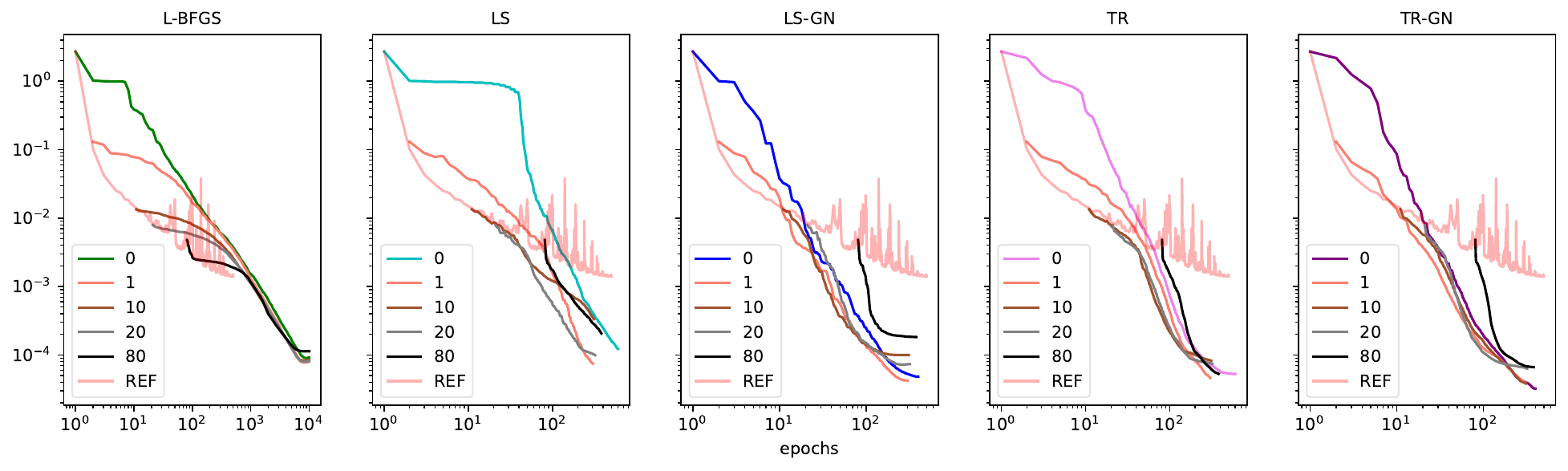}
\caption{FNO Burgers' test case with a hybrid solver. Convergence histories for testing metrics starting from checkpointed solutions of the reference solver. The values in the legends denote the checkpointed epoch. See Section \ref{sec:results_fno_burgers} for details.}
\label{fig:fno_conv_h}
\end{figure*}

We then consider a ``hybrid'' solver configuration, where the solvers are initialized using the parameters obtained by first performing an initial number of epochs using the reference solver; results are summarized in Figure \ref{fig:fno_conv_h}.
Performing a small number of epochs can allow further improvement in the generalization error in some cases, especially for LS and LS-GN; however, worse generalization performances can be observed if too many epochs are performed with the reference solver.

\subsubsection{Navier-Stokes equations} \label{sec:results_fno_ns}
In the second FNO test case, we consider the two-dimensional Navier-Stokes equations for a viscous, incompressible fluid in streamfunction-vorticity form as in Section 6.4 of \cite{FNO2},
\begin{equation}\label{eq:ns}
\begin{aligned}
\ut + \nabla^{\perp} \psi \cdot \nabla \u &= \nu \Delta \u + f,\\
-\Delta \psi &= \u,
\end{aligned}
\end{equation}
where $\nabla^\perp = (-\partial_y,\partial_x)$ is the rotated gradient and $\nu$ the viscosity. $\u$ and $\psi$ are the vorticity and the stream function of the fluid velocity field.
The spatial domain is $(0,1)^2$; the system is closed with periodic boundary conditions and with a given initial condition $\u(x,0) = \u_0(x)$.
Denoting by $\Psi$ the solution operator of Eq. \eqref{eq:ns}, here we are interested in learning the map
\begin{equation*}
\mathcal{G} : \Psi(\u_0,t)|_{t\in[0,10]} \to \Psi(\u_0,t)_{t\in[10,T]},
\end{equation*}
for a certain random distribution of the initial conditions $\u_0$ and a given final time $T > 10$.
The datasets are created by solving Eq. \eqref{eq:ns} with $\nu=0.001$ for Gaussian random initial conditions on a space-time grid $256 \times 256 \times 50000$ up to $T=50$ with a pseudo-spectral method, details are given in \cite{FNO}.
In our experiments, we used the same data used by the authors\footnote{Available at \link{drive.google.com/drive/folders/1UnbQh2WWc6knEHbLn-ZaXrKUZhp7pjt-}},
which is further subsampled to coarser grids: $64 \times 64 \times 10$ for the input, and $64 \times 64 \times 40$ for the output.
We use $1000$ input pairs for training and $200$ pairs as testing datasets to reproduce the experimental setting used to produce the results in the first column of Table 4 in \cite{FNO2}.
The network architecture is similar to the one used in Section \ref{sec:results_fno_burgers} using an additional normalization step for the input data. The output of each of the four Fourier layers is further processed by dense layers and summed to the output of a convolutional layer. The total number of parameters is $6,558,537$; the network is implemented reusing the original PyTorch code\footnote{Available at \link{github.com/neuraloperator/neuraloperator/blob/master/fourier\_3d.py}}.

\begin{figure}[h]
\centering
\includegraphics[width=\linewidth]{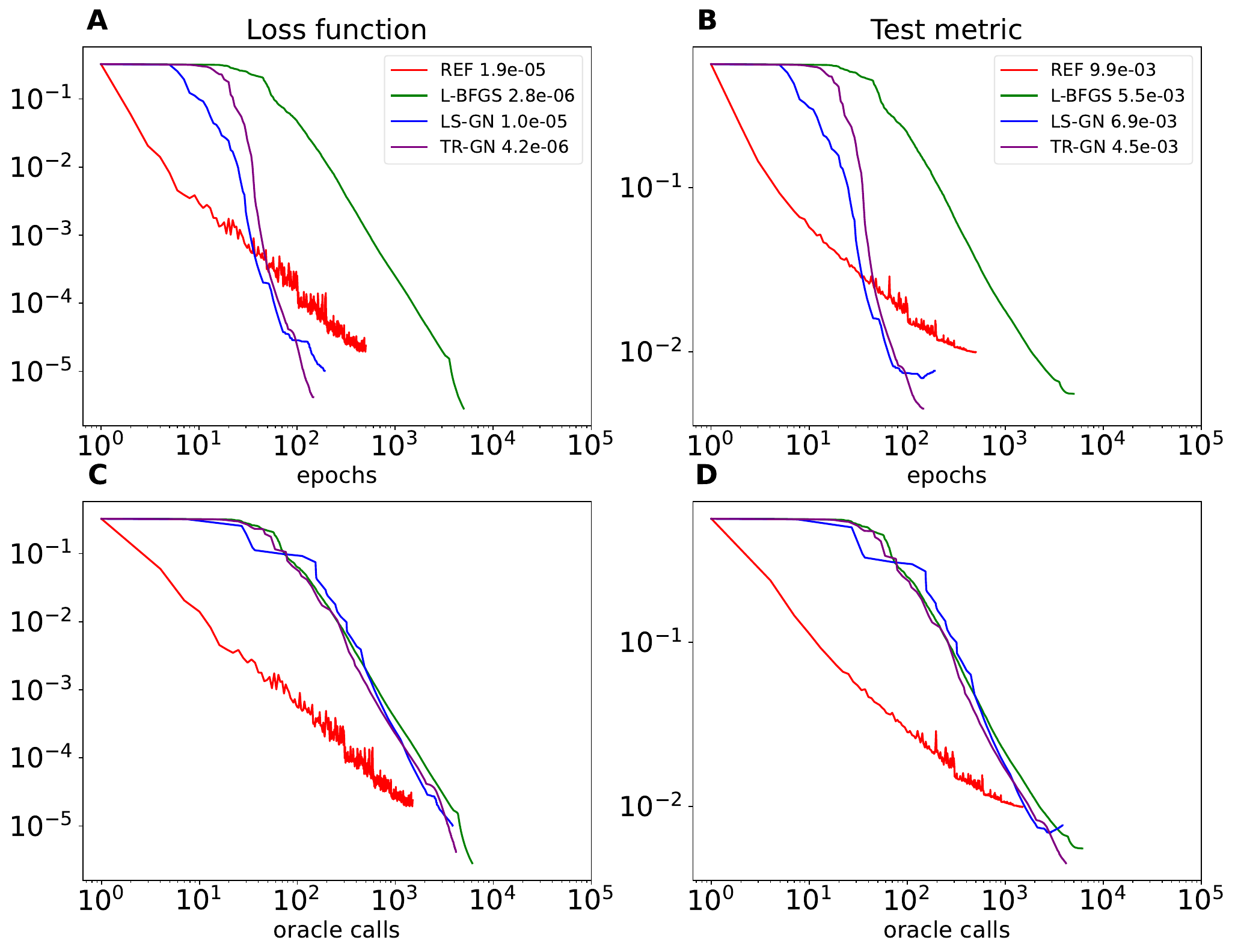}
\caption{FNO Navier-Stokes test case. Convergence histories for loss function values and testing metrics in terms of epochs (top row, panels A and B) and oracle calls (bottom row, panels C and D) for different solvers. The values in the legends denote the minimum metric value achieved. See Section \ref{sec:results_fno_ns} for details.}
\label{fig:fno_ns_conv}
\end{figure}

As before, the loss function is the mean squared $l2$ relative error as given in \eqref{ref:msrelerr}, while the accuracy metric is
the mean relative $l2$ error as given in \eqref{ref:mrelerr}.
The stochastic first-order solver used is the same as for the Burgers' test case, except that here the mini-batch size is 10.
Numerical results are reported in Fig. \ref{fig:fno_ns_conv}. We do not report the results for the exact Hessian variants of LS and TR since our PyTorch implementation of the matrix-vector product ran out of memory.

The reference solver shows a similar convergence history as for the Burgers' test case, reaching a final accuracy slightly larger than the one reported in \cite{FNO2} (0.0099 vs. 0.0089); the other solvers instead converge to local minima
characterized by smaller generalization errors, with TR-GN being able to further half the accuracy reported in \cite{FNO2}. Overfitting does not happen except for the LS-GN solver in the final stages of the optimization.
The costs of the solvers are comparable in terms of oracle calls, with TR-GN being the cheapest.

\subsection{DeepONet} \label{sec:results_deeponet}

Like the FNO technique briefly described in Section \ref{sec:results_fno}, DeepONet \cite{deeponet_natmachintell} tries to learn maps from input and output spaces. However, differently from FNO, the output of the operator maps an input function $u$ sampled at $m$ given ``sensors'' locations $\{x_i\}^m_{i=1}$ that do not need to be distributed on equispaced grids. The final goal is to approximate the map as
\begin{equation*}
\mathcal{G}_\p(u)(y) := \mathcal{B}(u(x_1),\dots,u(x_m)) \mathcal{T}(y)
\end{equation*}
where $\mathcal{B}$ and $\mathcal{T}$ are the ``branch'' and ``trunk'' networks and $y$ any point in the domain of $\mathcal{G}(u)$, not necessarily distributed at sensors locations.
For additional details, see \cite{deeponet_natmachintell}; for rigorous definitions, error estimates, and convergence rates, see \cite{DeepONet_mishra,DeepONet_schwab}.
A publicly available implementation of these techniques is provided in the Python package DeepXDE\footnote{Available at \link{github.com/lululxvi/deepxde}}.

\subsubsection{Reaction diffusion equations} \label{sec:results_deeponet_rd}
As a first test case, we consider the reaction-diffusion test case described as Problem 4 in \cite{deeponet_natmachintell}, where the goal is to learn an implicit operator associated with the one-dimensional nonlinear reaction-diffusion PDE
\begin{equation*}
\ut = D \uxx + k \u^2 + f, \quad x \in (0,1)
\end{equation*}
with homogeneous boundary conditions and zero initial conditions. $D$ is the diffusion coefficient while $k$ the reaction rate, both set to $0.01$ in the tests. Here, DeepONet is used to learn the operator mapping the forcing term $f(x)$ to the PDE solution $u(x,t)$ at $m$ random sensor locations.
The DeepONet architecture uses branch and trunk feed-forward networks of depth $3$, width $100$, and relu \cite{relu} activation functions for a total of $40,702$ parameters.
The network is implemented using the PyTorch backend and DeepXDE callbacks.

The data to be learned is generated using $N$ random forcing terms on a space-time grid $100 \times 100$; for each simulation, $P$ random points (different from the sensor locations) are selected. Here, we consider the configuration $N=1000$ for training, $N=10000$ for testing, with $P=100$ and $m=100$, which corresponds to the last data-point of Figure 4b in \cite{deeponet_natmachintell}.
We note that for this testing configuration, we were not able to find the data used to validate the original results; instead, we regenerate the data using the scripts provided by the authors\footnote{Available at \link{github.com/lululxvi/deeponet}}.
See the supplementary material in \cite{deeponet_natmachintell} for additional details.
The loss function and the accuracy metric are the usual mean squared error
\begin{equation}\label{ref:mse}
\frac{1}{N}\sum^N_{i=1}\|\y_i - \bary_i\|^2.
\end{equation}

\begin{figure}[h]
\centering
\includegraphics[width=\linewidth]{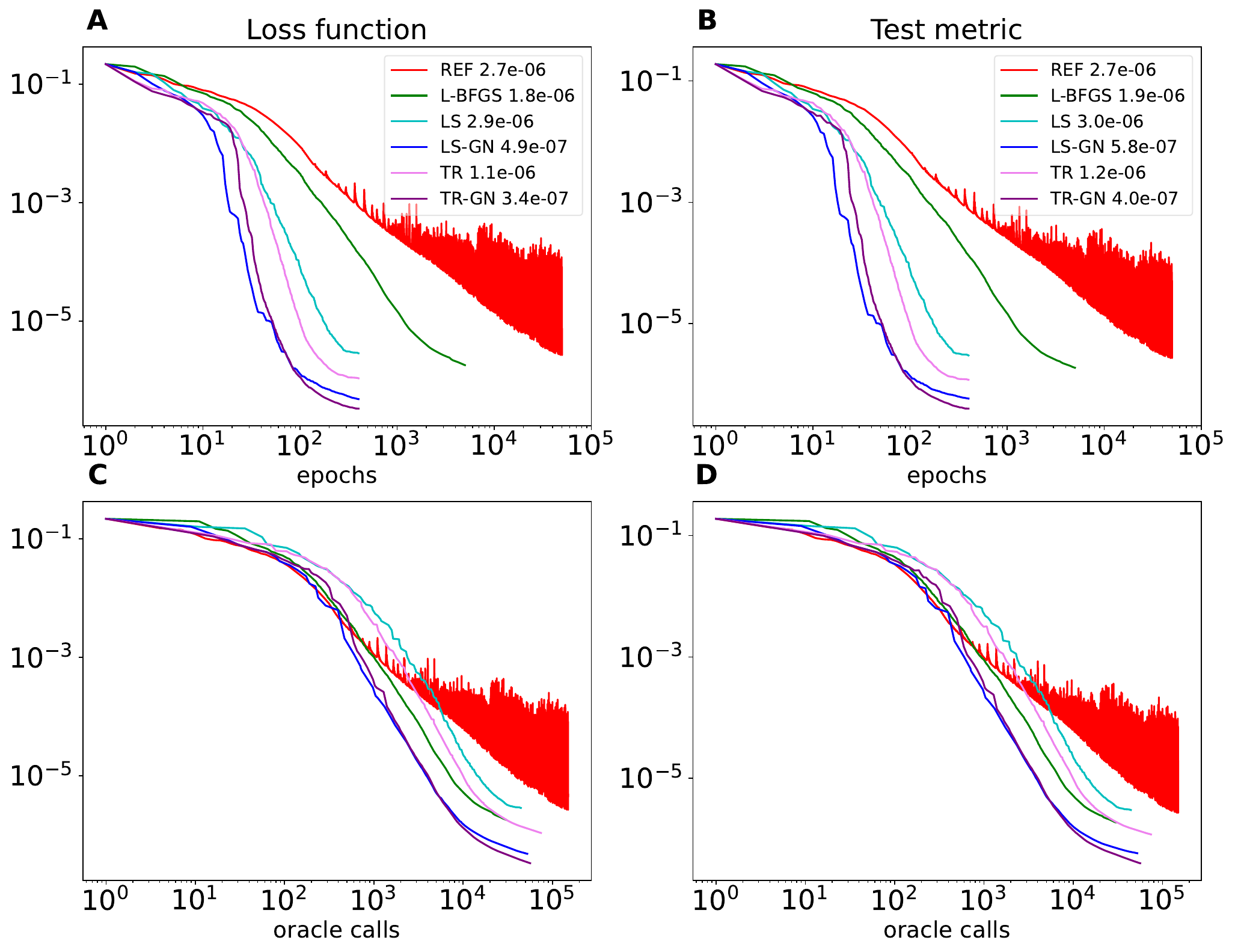}
\caption{DeepONet reaction-diffusion test case. Convergence histories for loss function values and testing metrics in terms of epochs (top row, panels A and B) and oracle calls (bottom row, panels C and D) for different solvers. The values in the legends denote the minimum metric value achieved. See Section \ref{sec:results_deeponet_rd} for details.}
\label{fig:deeponetrd_conv}
\end{figure}

The solver used in \cite{deeponet_natmachintell} is ADAM with a constant learning rate of $0.001$.
Numerical results are reported in Fig. \ref{fig:deeponetrd_conv}.

The reference solver, which is used deterministically here, converges more slowly than in the FNO test cases to a loss function value of $2.7$e-6, one order of magnitude smaller than the one reported in literature \cite{deeponet_natmachintell};
the achieved metric value is similar. Train and test errors for the reference solver are characterized by strong oscillations as the local minimum is approached.
The L-BFGS and LS solvers converge to loss and metric values similar to the reference solver, while the TR solver and the Gauss-Newton variants LS-GN and TR-GN are able to reduce the generalization errors further.
LS-GN exhibits a faster convergence rate in the early stages, followed by a phase of staircase convergence, while instead, the converge rate of TR-GN is more regular, leading to almost an order of magnitude improvement over the generalization error of the reference solver.
The computational costs of the solvers are similar, and they all require around $10^5$ oracle calls; like for the FNO test cases, the cheapest solver is TR-GN.

\begin{figure}[h]
\centering
\includegraphics[width=\linewidth]{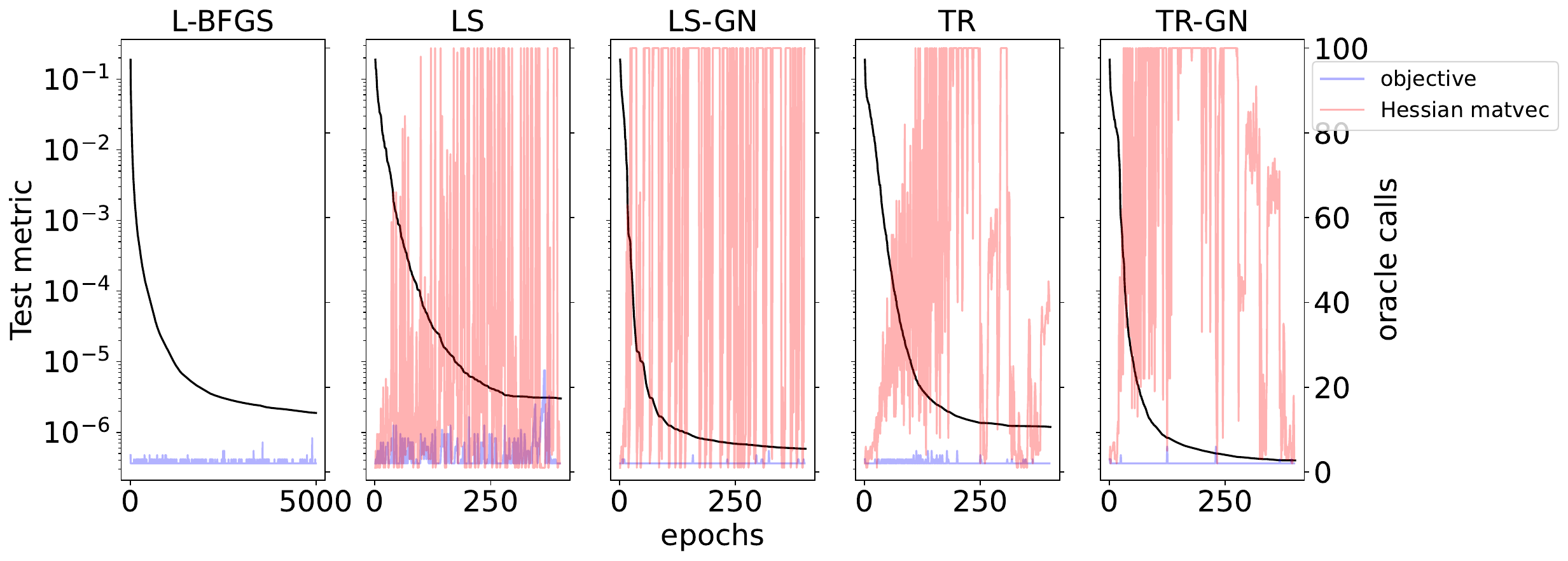}
\caption{DeepONet reaction-diffusion oracle calls breakdown. Test metric (black, left-most y-axis), objective function evaluations (shaded blue, right-most y-axis), and number of Hessian matrix-vector products (shaded red, right-most y-axis) against epoch number for different solvers (shown on top). See Section \ref{sec:results_deeponet_rd} for details.}
\label{fig:deeponetrd_cost}
\end{figure}
For this test case, we also provide a breakdown of the number of objective function evaluations and the Hessian matrix-vector products needed per epoch by the different solvers in Figure \ref{fig:deeponetrd_cost}.
Analyzing the number of objective function evaluations, L-BFGS on average requires 3 to 5 objective function calls per epoch, whereas LS-GN requires fewer.
On the other hand, LS requires many more objective evaluations; this indicates a struggle of the line-search procedure in identifying optimal descent, which leads to smaller update steps.
The trust region solvers instead utilize the objective function to decide on the acceptance of the model solution and eventually shrink the trust region;
the epochs in which the objective function is called more than once indicate the shrinking of the trust region.
On this aspect, the TR-GN solver is much more effective and requires the smallest number of objective function calls.
The major cost of the second-order solvers is in the number of Hessian matrix-vector multiplications, and both the trust region variants are very effective in the early stage of the optimization process,
with the number of matrix-vector products progressively ramping up to their limit (100) and then reducing in the final stages of the solvers.
The dynamic selection of the stopping criterion in the LS and LS-GN solver instead is not effective in controlling the number of linear iterations;
the detection of negative curvature further reduces the number of linear iterations in the early stages of the LS and TR solvers.

\subsubsection{Advection equations} \label{sec:results_deeponet_pod}

We then consider a variant of DeepONet where the trunk network is replaced by a precomputed basis for the output functions using the Proper Orthogonal Decomposition (POD). As a test case, we reproduce the testing configuration used in Section 5.4.1 in \cite{deeponet_fair}; specifically, we are interested in the one-dimensional advection equation
\begin{equation*}
\ut + \ux = 0, \quad x \in (0,1)
\end{equation*}
with periodic boundary conditions and learn the mapping from the randomly generated initial conditions $u_0(x)$ to the solution on the space-time domain $u(x,t)$, with $(x,t) \in [0,1]^2$.
The data is generated by solving the advection equation on a space-time grid of size $40 \times 40$ and considering two families of randomly generated initial conditions
\begin{align*}
\mbox{Case I: } &\, u_0(x) = h ~ \mathds{1}_{[c-w/2,c+w/2]},\\
\mbox{Case II: } &\, u_0(x) = h ~ \mathds{1}_{[c-w/2,c+w/2]} + \sqrt{\mbox{max}(d^2 - a^2(x-b)^2,0)},
\end{align*}
where $a, b, c, d, w, h$ are randomly selected (each in a separate interval) and $\mathds{1}$ is the indicator function. In both cases, the train and test datasets contain $1000$ pairs of random initialization and the corresponding space-time solution.
We use the data provided by the authors\footnote{Available at \link{github.com/lu-group/deeponet-fno}} as training and testing datasets for our computations; we also note that cases I/II here correspond to cases II/III in \cite{deeponet_fair} to avoid confusion.
The branch network is a feed-forward network with two layers of widths $512$ and $n$, where $n$ is the number of POD basis functions: for case I, $n = 38$, while for case II, $n=32$. The total number of parameters is $40,486$ for case I and $37,408$ for case II.
The networks are implemented using the PyTorch backend and DeepXDE callbacks.

\begin{figure}[H]
\centering
\includegraphics[width=\linewidth]{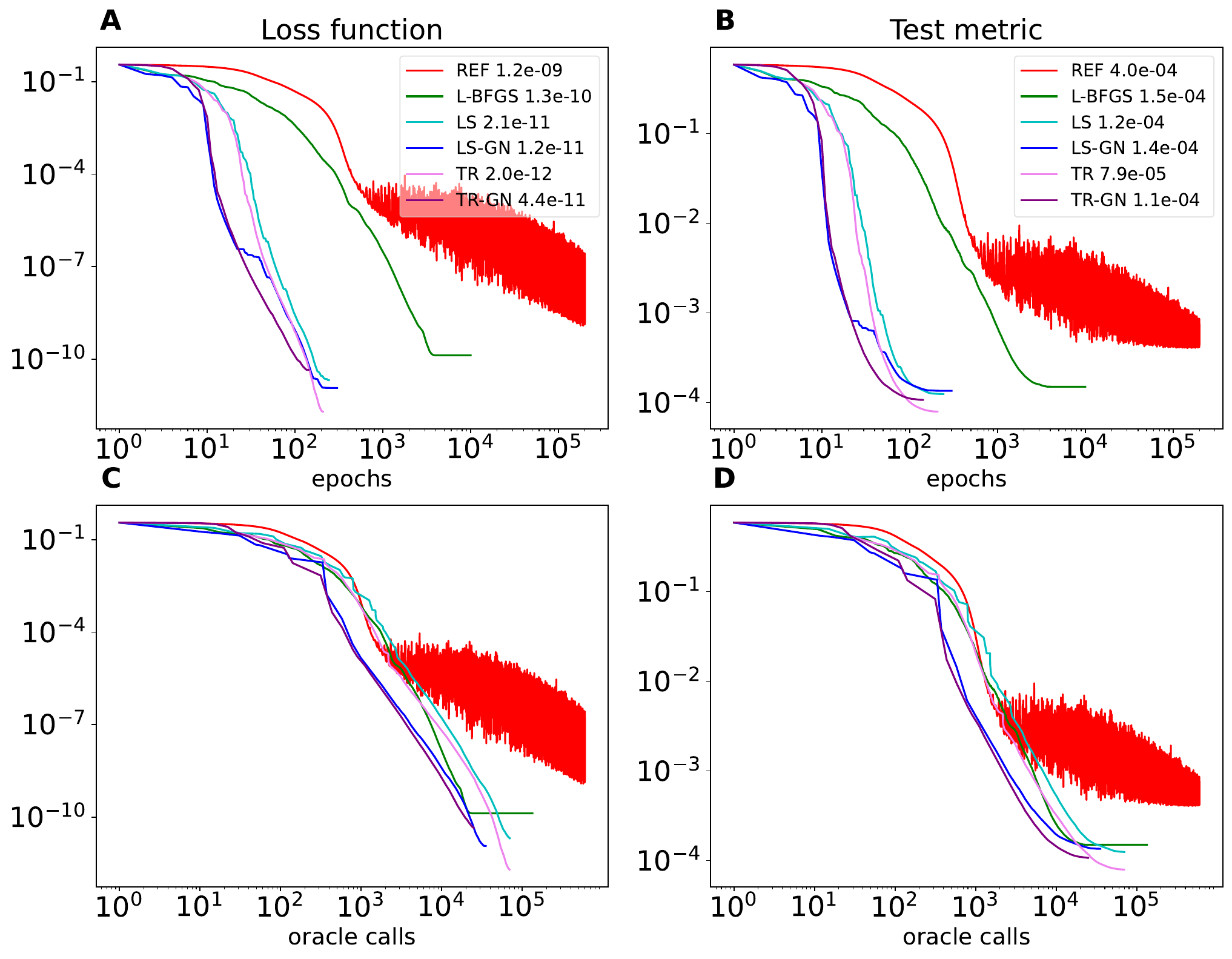}
\caption{DeepONet-POD test case I. Convergence histories for loss function values and testing metric in terms of epochs (top row, panels A and B) and oracle calls (bottom row, panels C and D) for different solvers. The values in the legends denote the minimum metric value achieved. See Section \ref{sec:results_deeponet_pod} for details.}
\label{fig:deeponet1_conv}
\end{figure}

\begin{figure}[H]
\centering
\includegraphics[width=\linewidth]{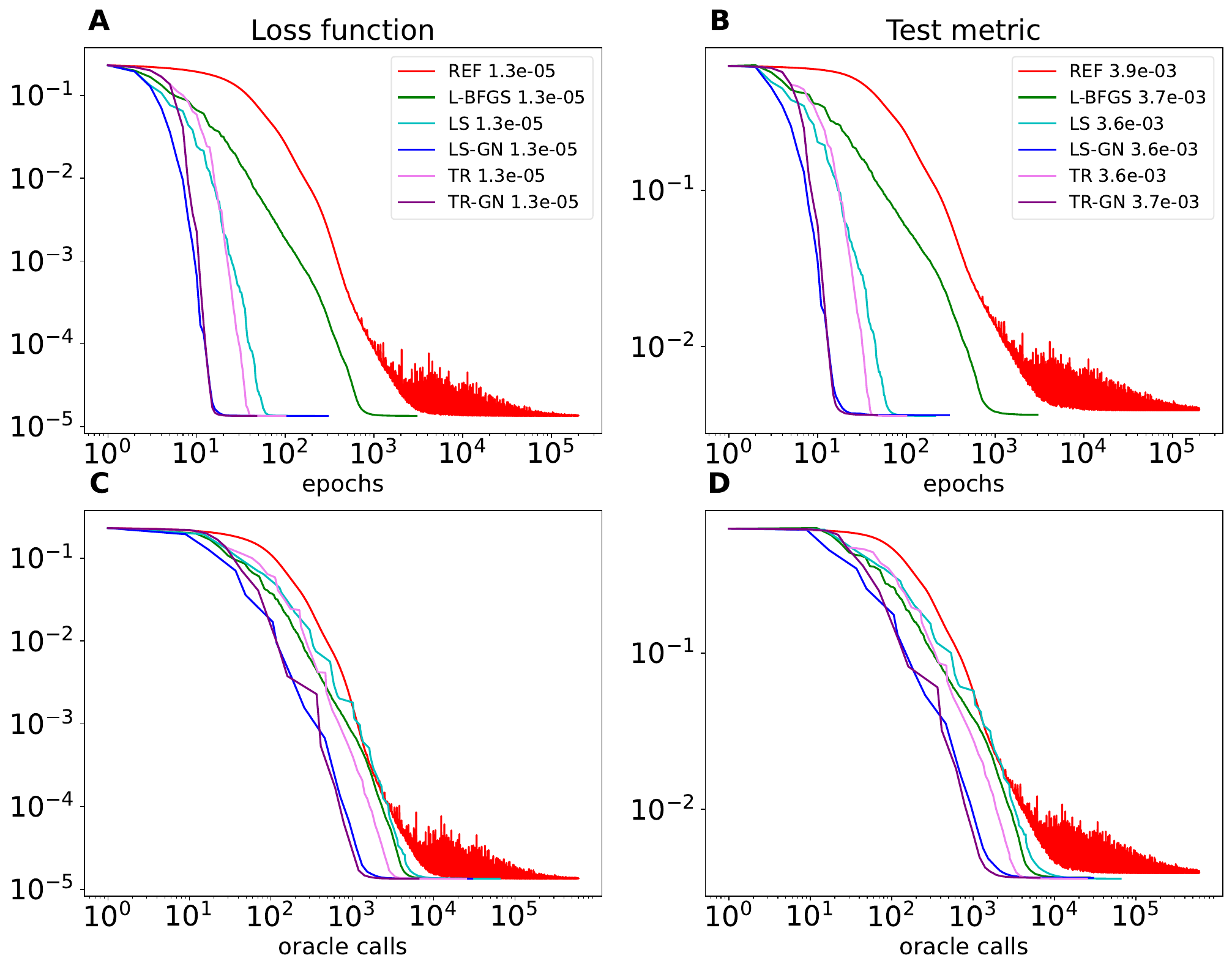}
\caption{DeepONet-POD test case II. Convergence histories for loss function values and testing metrics in terms of epochs (top row, panels A and B) and oracle calls (bottom row, panels C and D) for different solvers. The values in the legends denote the minimum metric value achieved. See Section \ref{sec:results_deeponet_pod} for details.}
\label{fig:deeponet2_conv}
\end{figure}

The loss function is the mean squared error \eqref{ref:mse}, and the mean squared $l2$ relative error given in \eqref{ref:msrelerr} is used as the accuracy metric.
The solver used in \cite{deeponet_fair} is ADAM with an initial learning rate of $0.001$, asymptotically decayed to $0.0001$ as the iteration progresses.
The computations are performed in single precision arithmetic as done in the accompanying software of \cite{deeponet_fair}; numerical results are reported in Fig. \ref{fig:deeponet1_conv} for case I and in Fig. \ref{fig:deeponet2_conv} for case II.

We first analyze the solvers' convergence histories for case I.
The reference solver exhibits a similar convergence history as the one in Section \ref{sec:results_deeponet_rd}; its generalization error is half of the one reported in Table 8 (Advection II column) in \cite{deeponet_fair}.
All the other solvers show faster convergence to different local minima characterized by smaller generalization errors, with the trust region variants achieving the smallest errors while still being faster than the other solvers in terms of oracle calls.

For test case II, all solvers converge to local minima possessing the same generalization capabilities. The error achieved by the reference solver is $0.0039$, perfectly aligned with the one reported in \cite{deeponet_fair} (see Table 8, Advection III column in the reference).
The major difference between solvers is in the number of oracle calls required to converge, with both the Gauss-Newton variants requiring a smaller number of calls.

\subsection{GreenLearning} \label{sec:results_greenlearning}

GreenLearning seeks to discover Green's functions associated with unknown linear operators by collecting physical system responses to random excitation inputs drawn from Gaussian processes \cite{greenlearning_nature}.
In a more formal setting, the goal is to learn the operator map by approximating the Green's function behind a physical process encoded by a linear operator
\begin{equation*}
L~u(x) = f(x), \quad x \in \Omega,
\end{equation*}
from a collection of input/output pairs $\{f_i, u_i\}$; the idea is to construct neural networks $\mathcal{N}_G$ and $\mathcal{N}_{h}$ such that
\begin{equation*}
u_i(x) \approx \int_\Omega \mathcal{N}_G(x,y) f_i(y) dy + \mathcal{N}_{h}(x)
\end{equation*}
where $\mathcal{N}_G$ approximates the Green's function and $\mathcal{N}_{h}$ a given homogeneous solution encoding the boundary conditions.

In this Section, we reproduce the experimental setting described in \cite{greenlearning} and consider the two-dimensional Poisson's equation on the unit square
\begin{equation*}
-\Delta \u = f
\end{equation*}
with zero boundary conditions, and learn the operator that maps forcing terms $f(x,t)$ to the solution of the PDE. The data is generated by solving the Poisson equation on a $421 \times 421$ grid with different forcing terms generated as Gaussian random fields and then subsampled on coarser grids of size $29 \times 29$. The training dataset used here contains $50$ elements, while the test dataset consists of $200$ elements, and they have been obtained from the data used in \cite{greenlearning}\footnote{Available at \link{zenodo.org/records/7701683}}.
The datasets are first normalized before being fed to the $\mathcal{N}_G$ network, characterized by a feed-forward architecture with depth $4$ and width $50$ that uses rational activation functions \cite{rational}. The current approximation of the Green's function, $\mathcal{N}_G$, is first sampled on a regular grid; the output obtained is then multiplied with the network input (i.e., an instance of the randomly generated forcing terms) mimicking the integration of Green's function. The total number of model parameters is $7,979$. The network is implemented using the PyTorch backend.

\begin{figure}[H]
\centering
\includegraphics[width=\linewidth]{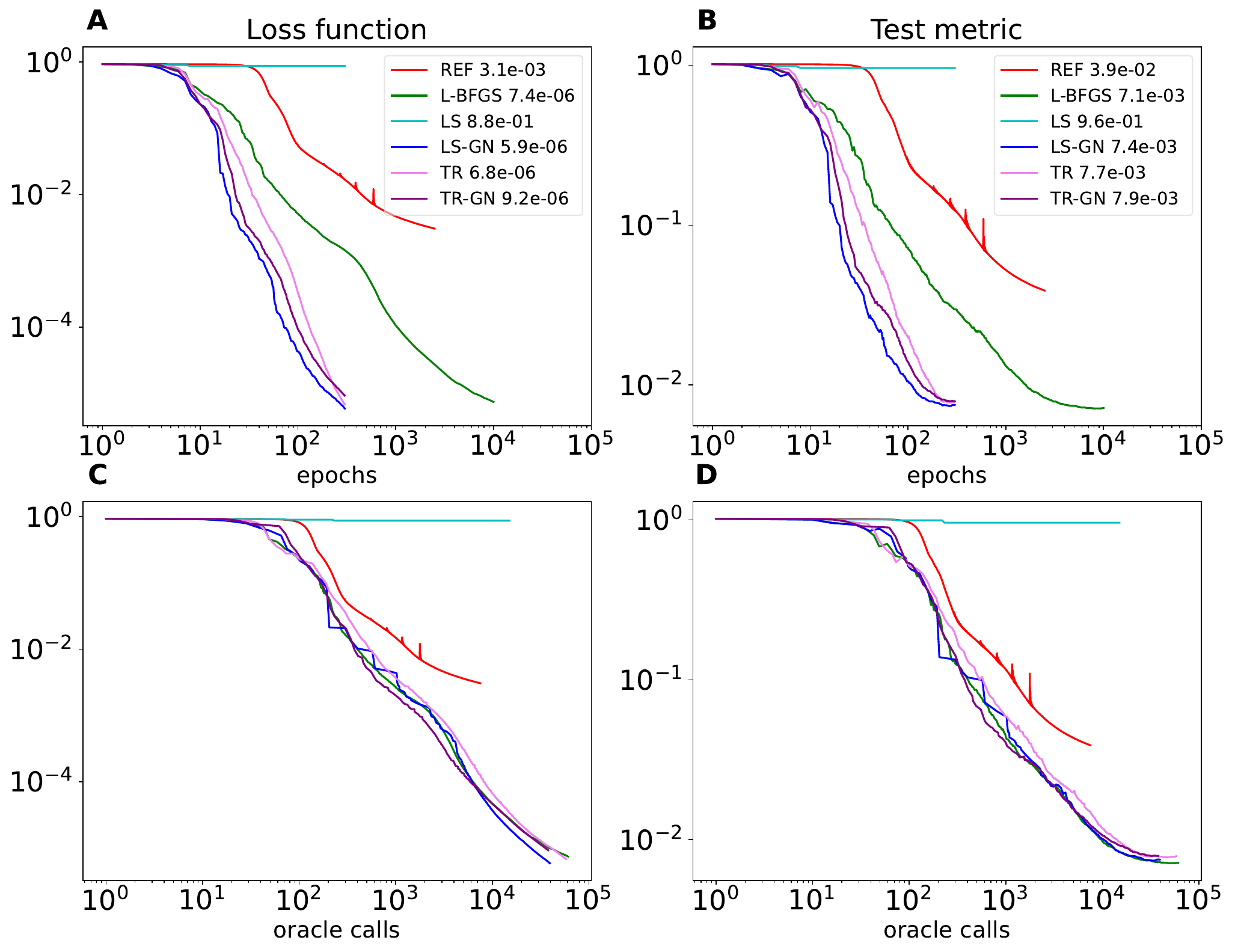}
\caption{GreenLearning test case. Convergence histories for loss function values and testing metrics in terms of epochs (top row, panels A and B) and oracle calls (bottom row, panels C and D) for different solvers. The values in the legends denote the minimum metric value achieved. See Section \ref{sec:results_greenlearning} for details.}
\label{fig:greenlearning_conv}
\end{figure}

The loss function and efficiency metric are the mean squared $l2$ relative error given in \eqref{ref:msrelerr}.
The solver used in \cite{greenlearning} is ADAMW with weight decay $0.0001$ and initial learning rate $0.001$, scaled by $0.9$ every 100 epochs using an exponential staircase schedule.
The computations are performed in single precision arithmetic as done in the accompanying software of \cite{greenlearning}, and numerical results are reported in Fig. \ref{fig:greenlearning_conv}; see Figure 1 in \cite{greenlearning} for the results with different training dataset sizes.

As for the other test cases reported in this Section, train and test metrics are strongly correlated, and no overfitting is observed.
The reference solver shows a very long plateau before converging with minimal oscillations; the achieved generalization error is aligned with the one reported in \cite{greenlearning}.
All the other solvers except for LS appear to converge to similar local minima characterized by generalization errors smaller than the reference case and with approximatively the same computational costs.
LS never escapes the initial plateau phase and fails to converge.

\section{Conclusions}\label{sec:conclusions}

In this paper, we have presented PETScML, a lightweight Python interface on top of PETSc, designed for solver experimentation in training deep-learning models.
Using PETScML, we have empirically demonstrated the efficacy of conventional second-order solvers in improving the generalization errors
arising from regression tasks when learning surrogate models, and with smaller computational costs.

Among the various solvers tested, the trust region method with Gauss-Newton approximation of the Hessian proved to be the overall best-performing method for the test cases analyzed,
either in terms of final model accuracy or in terms of computational costs.

Much remains to follow up on these encouraging results, in particular on the design of linear preconditioning techniques for limiting
the number of Krylov iterations, and on the design of nonlinear preconditioning strategies \cite{cai2002nonlinearly,brune2015composing,liu2024overlapping} to limit the number of nonlinear iterations and improve the overall convergence rate of the solver.
The convergence properties of the trust region method can also be further improved by using different trust region norms and ellipsoidal scaling techniques.
These topics will be the subject of future work.

Other SciML techniques based on regression tasks can potentially benefit from the adoption of second-order solvers;
possible future directions in this context include molecular dynamics simulations \cite{wen2022deep,gordonbell23} and continuous heterogeneous cryo-EM reconstructions \cite{cryodrgn}.

Future work will also focus on the application of second-order solvers to more conventional deep-learning tasks like image classification \cite{densenet}, sequence modeling \cite{irie2021going}, vision tasks \cite{dosovitskiy2020image}, and self-supervised learning \cite{elmo},
where we plan to study the ability of second-order methods to reach state-of-the-art validation accuracies with minor hyper-parameter tuning,
ideally in less wall-clock time or energy costs.

\section*{Acknowledgments}
Umberto Zerbinati acknowledges the support of KAUST, since part of the research was conducted when he was a student in MSc program of the CEMSE division.
The authors acknowledge the use of the Ibex GPU Cluster of the KAUST Supercomputing Laboratory and support from the Extreme Computing Research Center and the AI Initiative at KAUST.

\enlargethispage*{0.5in}
\clearpage
\bibliographystyle{plain}
\bibliography{paper}

\begin{thebibliography}{10}

\bibitem{EW4}
Heng-Bin An, Ze-Yao Mo, and Xing-Ping Liu.
\newblock A choice of forcing terms in inexact {N}ewton method.
\newblock {\em Journal of Computational and Applied Mathematics},
  200(1):47--60, 2007.

\bibitem{anil20}
Rohan Anil, Vineet Gupta, Tomer Koren, Kevin Regan, and Yoram Singer.
\newblock Second order optimization made practical.
\newblock {\em CoRR}, abs/2002.09018, 2020.

\bibitem{petsc}
Satish Balay, Shrirang Abhyankar, Mark~F. Adams, Steven Benson, Jed Brown,
  Peter Brune, Kris Buschelman, Emil Constantinescu, Lisandro Dalcin, Alp
  Dener, Victor Eijkhout, Jacob Faibussowitsch, William~D. Gropp, V\'{a}clav
  Hapla, Tobin Isaac, Pierre Jolivet, Dmitry Karpeev, Dinesh Kaushik,
  Matthew~G. Knepley, Fande Kong, Scott Kruger, Dave~A. May, Lois~Curfman
  McInnes, Richard~Tran Mills, Lawrence Mitchell, Todd Munson, Jose~E. Roman,
  Karl Rupp, Patrick Sanan, Jason Sarich, Barry~F. Smith, Stefano Zampini, Hong
  Zhang, Hong Zhang, and Junchao Zhang.
\newblock {PETSc/TAO} users manual.
\newblock Technical Report ANL-21/39 - Revision 3.20, Argonne National
  Laboratory, 2023.

\bibitem{petsc-web-page}
Satish Balay, Shrirang Abhyankar, Mark~F. Adams, Steven Benson, Jed Brown,
  Peter Brune, Kris Buschelman, Emil~M. Constantinescu, Lisandro Dalcin, Alp
  Dener, Victor Eijkhout, Jacob Faibussowitsch, William~D. Gropp, V\'{a}clav
  Hapla, Tobin Isaac, Pierre Jolivet, Dmitry Karpeev, Dinesh Kaushik,
  Matthew~G. Knepley, Fande Kong, Scott Kruger, Dave~A. May, Lois~Curfman
  McInnes, Richard~Tran Mills, Lawrence Mitchell, Todd Munson, Jose~E. Roman,
  Karl Rupp, Patrick Sanan, Jason Sarich, Barry~F. Smith, Stefano Zampini, Hong
  Zhang, Hong Zhang, and Junchao Zhang.
\newblock {PETS}c {W}eb page, 2023.

\bibitem{BFGSpre}
Luca Bergamaschi, Jos{\'e} Mar{\'\i}n, and {\'A}ngeles Mart{\'\i}nez.
\newblock Compact quasi-{N}ewton preconditioners for symmetric positive
  definite linear systems.
\newblock {\em Numerical Linear Algebra with Applications}, 27(6):e2322, 2020.

\bibitem{nocedal18b}
L{\'e}on Bottou, Frank~E. Curtis, and Jorge Nocedal.
\newblock Optimization methods for large-scale machine learning.
\newblock {\em SIAM Review}, 60(2):223--311, 2018.

\bibitem{greenlearning_nature}
Nicolas Boull{\'e}, Christopher~J Earls, and Alex Townsend.
\newblock Data-driven discovery of {G}reen's functions with
  human-understandable deep learning.
\newblock {\em Scientific reports}, 12(1):4824, 2022.

\bibitem{greenlearning}
Nicolas Boull{\'e}, Diana Halikias, and Alex Townsend.
\newblock Elliptic {PDE} learning is provably data-efficient.
\newblock {\em Proceedings of the National Academy of Sciences},
  120(39):e2303904120, 2023.

\bibitem{rational}
Nicolas Boull{\'e}, Yuji Nakatsukasa, and Alex Townsend.
\newblock Rational neural networks.
\newblock {\em Advances in neural information processing systems},
  33:14243--14253, 2020.

\bibitem{brune2015composing}
Peter~R Brune, Matthew~G Knepley, Barry~F Smith, and Xuemin Tu.
\newblock Composing scalable nonlinear algebraic solvers.
\newblock {\em SIAM Review}, 57(4):535--565, 2015.

\bibitem{brunton2016discovering}
Steven~L Brunton, Joshua~L Proctor, and J~Nathan Kutz.
\newblock Discovering governing equations from data by sparse identification of
  nonlinear dynamical systems.
\newblock {\em Proceedings of the national academy of sciences},
  113(15):3932--3937, 2016.

\bibitem{BFGSs}
Richard~H Byrd, Samantha~L Hansen, Jorge Nocedal, and Yoram Singer.
\newblock A stochastic quasi-{N}ewton method for large-scale optimization.
\newblock {\em SIAM Journal on Optimization}, 26(2):1008--1031, 2016.

\bibitem{denseQN}
Richard~H Byrd, Jorge Nocedal, and Robert~B Schnabel.
\newblock Representations of quasi-{N}ewton matrices and their use in limited
  memory methods.
\newblock {\em Mathematical Programming}, 63(1-3):129--156, 1994.

\bibitem{cai2002nonlinearly}
Xiao-Chuan Cai and David~E Keyes.
\newblock Nonlinearly preconditioned inexact newton algorithms.
\newblock {\em SIAM Journal on Scientific Computing}, 24(1):183--200, 2002.

\bibitem{ConnGould}
Andrew~R Conn, Nicholas~IM Gould, and Philippe~L Toint.
\newblock {\em Trust region methods}.
\newblock SIAM, 2000.

\bibitem{cybenko1989}
George Cybenko.
\newblock Approximation by superpositions of a sigmoidal function.
\newblock {\em Mathematics of control, signals and systems}, 2(4):303--314,
  1989.

\bibitem{petsc4py}
Lisandro~D Dalcin, Rodrigo~R Paz, Pablo~A Kler, and Alejandro Cosimo.
\newblock Parallel distributed computing using {P}ython.
\newblock {\em Advances in Water Resources}, 34(9):1124--1139, 2011.

\bibitem{gordonbell23}
Sambit Das, Bikash Kanungo, Vishal Subramanian, Gourab Panigrahi, Phani
  Motamarri, David Rogers, Paul Zimmerman, and Vikram Gavini.
\newblock Large-scale materials modeling at quantum accuracy: Ab initio
  simulations of quasicrystals and interacting extended defects in metallic
  alloys.
\newblock In {\em Proceedings of the International Conference for High
  Performance Computing, Networking, Storage and Analysis}, SC '23, New York,
  NY, USA, 2023. Association for Computing Machinery.

\bibitem{de2022cost}
Maarten~V de~Hoop, Daniel~Zhengyu Huang, Elizabeth Qian, and Andrew~M Stuart.
\newblock The cost-accuracy trade-off in operator learning with neural
  networks.
\newblock {\em Journal of Machine Learning}, 2022.

\bibitem{dosovitskiy2020image}
Alexey Dosovitskiy, Lucas Beyer, Alexander Kolesnikov, Dirk Weissenborn,
  Xiaohua Zhai, Thomas Unterthiner, Mostafa Dehghani, Matthias Minderer, Georg
  Heigold, Sylvain Gelly, et~al.
\newblock An image is worth 16x16 words: Transformers for image recognition at
  scale.
\newblock {\em arXiv preprint arXiv:2010.11929}, 2020.

\bibitem{ADAGRAD}
John Duchi, Elad Hazan, and Yoram Singer.
\newblock Adaptive subgradient methods for online learning and stochastic
  optimization.
\newblock {\em Journal of machine learning research}, 12(7), 2011.

\bibitem{EisenstatWalker}
Stanley~C Eisenstat and Homer~F Walker.
\newblock Choosing the forcing terms in an inexact {N}ewton method.
\newblock {\em SIAM Journal on Scientific Computing}, 17(1):16--32, 1996.

\bibitem{jax}
Roy Frostig, Matthew~James Johnson, and Chris Leary.
\newblock Compiling machine learning programs via high-level tracing.
\newblock {\em Systems for Machine Learning}, 4(9), 2018.

\bibitem{goldfarb20}
Donald Goldfarb, Yi~Ren, and Achraf Bahamou.
\newblock Practical quasi-newton methods for training deep neural networks.
\newblock In {\em Proceedings of the 34th International Conference on Neural
  Information Processing Systems}, NeurIPS, pages 201:1--11, 2020.

\bibitem{goyal2017accurate}
Priya Goyal, Piotr Doll{\'a}r, Ross Girshick, Pieter Noordhuis, Lukasz
  Wesolowski, Aapo Kyrola, Andrew Tulloch, Yangqing Jia, and Kaiming He.
\newblock Accurate, large minibatch sgd: Training imagenet in 1 hour.
\newblock {\em arXiv preprint arXiv:1706.02677}, 2017.

\bibitem{shampoo}
Vineet Gupta, Tomer Koren, and Yoram Singer.
\newblock Shampoo: Preconditioned stochastic tensor optimization.
\newblock 2018.

\bibitem{GELU}
Dan Hendrycks and Kevin Gimpel.
\newblock Gaussian {E}rror {L}inear {U}nits ({GELU}s).
\newblock {\em arXiv preprint arXiv:1606.08415}, 2016.

\bibitem{hornik1989multilayer}
Kurt Hornik, Maxwell Stinchcombe, and Halbert White.
\newblock Multilayer feedforward networks are universal approximators.
\newblock {\em Neural networks}, 2(5):359--366, 1989.

\bibitem{densenet}
Gao Huang, Zhuang Liu, Laurens Van Der~Maaten, and Kilian~Q Weinberger.
\newblock Densely connected convolutional networks.
\newblock In {\em Proceedings of the IEEE conference on computer vision and
  pattern recognition}, pages 4700--4708, 2017.

\bibitem{irie2021going}
Kazuki Irie, Imanol Schlag, R{\'o}bert Csord{\'a}s, and J{\"u}rgen Schmidhuber.
\newblock Going beyond linear transformers with recurrent fast weight
  programmers.
\newblock {\em Advances in Neural Information Processing Systems},
  34:7703--7717, 2021.

\bibitem{jia2020pushing}
Weile Jia, Han Wang, Mohan Chen, Denghui Lu, Lin Lin, Roberto Car, E~Weinan,
  and Linfeng Zhang.
\newblock Pushing the limit of molecular dynamics with ab initio accuracy to
  100 million atoms with machine learning.
\newblock In {\em SC20: International conference for high performance
  computing, networking, storage and analysis}, pages 1--14. IEEE, 2020.

\bibitem{ADAM}
Diederik~P Kingma and Jimmy Ba.
\newblock {ADAM}: A method for stochastic optimization.
\newblock {\em arXiv preprint arXiv:1412.6980}, 2014.

\bibitem{FNOuniversal}
Nikola Kovachki, Samuel Lanthaler, and Siddhartha Mishra.
\newblock On universal approximation and error bounds for fourier neural
  operators.
\newblock {\em The Journal of Machine Learning Research}, 22(1):13237--13312,
  2021.

\bibitem{FNO2}
Nikola~B Kovachki, Zongyi Li, Burigede Liu, Kamyar Azizzadenesheli, Kaushik
  Bhattacharya, Andrew~M Stuart, and Anima Anandkumar.
\newblock Neural {O}perator: Learning maps between function spaces with
  applications to {PDE}s.
\newblock {\em J. Mach. Learn. Res.}, 24(89):1--97, 2023.

\bibitem{DeepONet_mishra}
Samuel Lanthaler, Siddhartha Mishra, and George~E Karniadakis.
\newblock {Error estimates for {D}eep{ON}ets: a deep learning framework in
  infinite dimensions}.
\newblock {\em Transactions of Mathematics and Its Applications}, 6(1):tnac001,
  03 2022.

\bibitem{lecun2015deep}
Yann LeCun, Yoshua Bengio, and Geoffrey Hinton.
\newblock Deep learning.
\newblock {\em nature}, 521(7553):436--444, 2015.

\bibitem{PSGD}
Xi-Lin Li.
\newblock Preconditioned stochastic gradient descent.
\newblock {\em IEEE transactions on neural networks and learning systems},
  29(5):1454--1466, 2017.

\bibitem{FNO}
Zongyi Li, Nikola Kovachki, Kamyar Azizzadenesheli, Burigede Liu, Kaushik
  Bhattacharya, Andrew Stuart, and Anima Anandkumar.
\newblock {F}ourier {N}eural {O}perator for parametric partial differential
  equations.
\newblock {\em arXiv preprint arXiv:2010.08895}, 2020.

\bibitem{lbfgs}
Dong~C Liu and Jorge Nocedal.
\newblock On the limited memory bfgs method for large scale optimization.
\newblock {\em Mathematical programming}, 45(1-3):503--528, 1989.

\bibitem{liu2024overlapping}
Lulu Liu, Weiguo Gao, Han Yu, and David~E Keyes.
\newblock Overlapping multiplicative schwarz preconditioning for linear and
  nonlinear systems.
\newblock {\em Journal of Computational Physics}, 496:112548, 2024.

\bibitem{newtonMR}
Yang Liu and Fred Roosta.
\newblock Convergence of {Newton-MR} under inexact {H}essian information.
\newblock {\em SIAM Journal on Optimization}, 31(1):59--90, 2021.

\bibitem{ADAMW}
Ilya Loshchilov and Frank Hutter.
\newblock Decoupled weight decay regularization.
\newblock {\em arXiv preprint arXiv:1711.05101}, 2017.

\bibitem{deeponet_natmachintell}
Lu~Lu, Pengzhan Jin, Guofei Pang, Zhongqiang Zhang, and George~Em Karniadakis.
\newblock Learning nonlinear operators via {D}eep{ON}et based on the universal
  approximation theorem of operators.
\newblock {\em Nature machine intelligence}, 3(3):218--229, 2021.

\bibitem{deeponet_fair}
Lu~Lu, Xuhui Meng, Shengze Cai, Zhiping Mao, Somdatta Goswami, Zhongqiang
  Zhang, and George~Em Karniadakis.
\newblock A comprehensive and fair comparison of two neural operators (with
  practical extensions) based on fair data.
\newblock {\em Computer Methods in Applied Mechanics and Engineering},
  393:114778, 2022.

\bibitem{DeepONet_schwab}
Carlo Marcati and Christoph Schwab.
\newblock Exponential convergence of deep operator networks for elliptic
  partial differential equations.
\newblock {\em SIAM Journal on Numerical Analysis}, 61(3):1513--1545, 2023.

\bibitem{KFAC}
James Martens and Roger Grosse.
\newblock Optimizing neural networks with kronecker-factored approximate
  curvature.
\newblock In {\em International conference on machine learning}, pages
  2408--2417. PMLR, 2015.

\bibitem{martensGN}
James Martens and Ilya Sutskever.
\newblock Learning recurrent neural networks with {H}essian-free optimization.
\newblock In {\em Proceedings of the 28th international conference on machine
  learning (ICML-11)}, pages 1033--1040, 2011.

\bibitem{petsc-gpu}
Richard~Tran Mills, Mark~F. Adams, Satish Balay, Jed Brown, Alp Dener, Matthew
  Knepley, Scott~E. Kruger, Hannah Morgan, Todd Munson, Karl Rupp, Barry~F.
  Smith, Stefano Zampini, Hong Zhang, and Junchao Zhang.
\newblock Toward performance-portable {PETS}c for {GPU}-based exascale systems.
\newblock {\em Parallel Computing}, 108:102831, 2021.

\bibitem{relu}
Vinod Nair and Geoffrey~E Hinton.
\newblock Rectified linear units improve restricted boltzmann machines.
\newblock In {\em Proceedings of the 27th international conference on machine
  learning (ICML-10)}, pages 807--814, 2010.

\bibitem{nakamura2021adaptive}
Tenavi Nakamura-Zimmerer, Qi~Gong, and Wei Kang.
\newblock Adaptive deep learning for high-dimensional hamilton--jacobi--bellman
  equations.
\newblock {\em SIAM Journal on Scientific Computing}, 43(2):A1221--A1247, 2021.

\bibitem{NocedalAndWright}
Jorge Nocedal and Stephen~J Wright.
\newblock {\em Numerical optimization}.
\newblock Springer, 2006.

\bibitem{olearyroseberry21}
Thomas O'Leary-Roseberry, Nick Alger, and Omar Ghattas.
\newblock Low rank saddle free newton: A scalable method for stochastic
  nonconvex optimization, 2021.

\bibitem{dino}
Thomas O'Leary-Roseberry, Peng Chen, Umberto Villa, and Omar Ghattas.
\newblock Derivative-informed neural operator: An efficient framework for
  high-dimensional parametric derivative learning.
\newblock {\em Journal of Computational Physics}, 496:112555, 2024.

\bibitem{ASDL}
Kazuki Osawa, Satoki Ishikawa, Rio Yokota, Shigang Li, and Torsten Hoefler.
\newblock {ASDL}: A unified interface for gradient preconditioning in pytorch.
\newblock {\em arXiv preprint arXiv:2305.04684}, 2023.

\bibitem{rio18}
Kazuki Osawa, Yohei Tsuji, Yuichiro Ueno, Akira Naruse, Rio Yokota, and
  S.~Matsuoka.
\newblock Second-order optimization method for large mini-batch: Training
  resnet-50 on imagenet in 35 epochs.
\newblock {\em ArXiv}, abs/1811.12019, 2018.

\bibitem{pytorch}
Adam Paszke, Sam Gross, Francisco Massa, Adam Lerer, James Bradbury, Gregory
  Chanan, Trevor Killeen, Zeming Lin, Natalia Gimelshein, Luca Antiga, Alban
  Desmaison, Andreas Kopf, Edward Yang, Zachary DeVito, Martin Raison, Alykhan
  Tejani, Sasank Chilamkurthy, Benoit Steiner, Lu~Fang, Junjie Bai, and Soumith
  Chintala.
\newblock Pytorch: An imperative style, high-performance deep learning library.
\newblock {\em Advances in neural information processing systems}, 32, 2019.

\bibitem{kaisa}
J.~Gregory Pauloski, Qi~Huang, Lei Huang, Shivaram Venkataraman, Kyle Chard,
  Ian Foster, and Zhao Zhang.
\newblock {KAISA}: {A}n adaptive second-order optimizer framework for deep
  neural networks.
\newblock In {\em Proceedings of the International Conference for High
  Performance Computing, Networking, Storage and Analysis}, SC '21, New York,
  NY, USA, 2021. Association for Computing Machinery.

\bibitem{raissi2018deep}
Maziar Raissi.
\newblock Deep hidden physics models: Deep learning of nonlinear partial
  differential equations.
\newblock {\em The Journal of Machine Learning Research}, 19(1):932--955, 2018.

\bibitem{raissi2019physics}
Maziar Raissi, Paris Perdikaris, and George~E Karniadakis.
\newblock Physics-informed neural networks: A deep learning framework for
  solving forward and inverse problems involving nonlinear partial differential
  equations.
\newblock {\em Journal of Computational physics}, 378:686--707, 2019.

\bibitem{LSs}
Farbod Roosta-Khorasani and Michael~W Mahoney.
\newblock Sub-sampled newton methods.
\newblock {\em Mathematical Programming}, 174:293--326, 2019.

\bibitem{elmo}
Justyna Sarzynska-Wawer, Aleksander Wawer, Aleksandra Pawlak, Julia
  Szymanowska, Izabela Stefaniak, Michal Jarkiewicz, and Lukasz Okruszek.
\newblock Detecting formal thought disorder by deep contextualized word
  representations.
\newblock {\em Psychiatry Research}, 304:114135, 2021.

\bibitem{schmidhuber2015deep}
J{\"u}rgen Schmidhuber.
\newblock Deep learning in neural networks: An overview.
\newblock {\em Neural networks}, 61:85--117, 2015.

\bibitem{schraudolph}
Nicol~N Schraudolph.
\newblock Fast curvature matrix-vector products for second-order gradient
  descent.
\newblock {\em Neural computation}, 14(7):1723--1738, 2002.

\bibitem{sgd_stoch_reg}
Samuel~L Smith, Benoit Dherin, David~GT Barrett, and Soham De.
\newblock On the origin of implicit regularization in stochastic gradient
  descent.
\newblock {\em 9th International Conference on Learning Representations, ICLR
  2021}, 2021.

\bibitem{Steihaug}
Trond Steihaug.
\newblock The conjugate gradient method and trust regions in large scale
  optimization.
\newblock {\em SIAM Journal on Numerical Analysis}, 20(3):626--637, 1983.

\bibitem{momentumSGD}
Ilya Sutskever, James Martens, George Dahl, and Geoffrey Hinton.
\newblock On the importance of initialization and momentum in deep learning.
\newblock In {\em International conference on machine learning}, pages
  1139--1147. PMLR, 2013.

\bibitem{Toint}
Philippe Toint.
\newblock Towards an efficient sparsity exploiting newton method for
  minimization.
\newblock In {\em Sparse matrices and their uses}, pages 57--88. Academic
  press, 1981.

\bibitem{vandenbrand21}
Jan van~den Brand, Binghui Peng, Zhao Song, and Omri Weinstein.
\newblock {Training (Overparametrized) Neural Networks in Near-Linear Time}.
\newblock In James~R. Lee, editor, {\em 12th Innovations in Theoretical
  Computer Science Conference (ITCS 2021)}, volume 185 of {\em Leibniz
  International Proceedings in Informatics (LIPIcs)}, pages 63:1--63:15,
  Dagstuhl, Germany, 2021. Schloss Dagstuhl--Leibniz-Zentrum f{\"u}r
  Informatik.

\bibitem{wen2022deep}
Tongqi Wen, Linfeng Zhang, Han Wang, E~Weinan, and David~J Srolovitz.
\newblock Deep potentials for materials science.
\newblock {\em Materials Futures}, 1(2):022601, 2022.

\bibitem{TRs}
Peng Xu, Fred Roosta, and Michael~W Mahoney.
\newblock Newton-type methods for non-convex optimization under inexact
  {H}essian information.
\newblock {\em Mathematical Programming}, 184(1-2):35--70, 2020.

\bibitem{SENG}
Minghan Yang, Dong Xu, Zaiwen Wen, Mengyun Chen, and Pengxiang Xu.
\newblock Sketch-based empirical natural gradient methods for deep learning.
\newblock {\em Journal of Scientific Computing}, 92(3):94, 2022.

\bibitem{adahessian}
Zhewei Yao, Amir Gholami, Sheng Shen, Mustafa Mustafa, Kurt Keutzer, and
  Michael Mahoney.
\newblock {AdaHessian}: An adaptive second order optimizer for machine
  learning.
\newblock {\em Proceedings of the AAAI Conference on Artificial Intelligence},
  35(12):10665--10673, May 2021.

\bibitem{lars}
Yang You, Igor Gitman, and Boris Ginsburg.
\newblock Scaling sgd batch size to 32k for imagenet training.
\newblock {\em arXiv preprint arXiv:1708.03888}, 6(12):6, 2017.

\bibitem{lamb}
Yang You, Jing Li, Sashank Reddi, Jonathan Hseu, Sanjiv Kumar, Srinadh
  Bhojanapalli, Xiaodan Song, James Demmel, Kurt Keutzer, and Cho-Jui Hsieh.
\newblock Large batch optimization for deep learning: training bert in 76
  minutes.
\newblock 2020.

\bibitem{yu2018deep}
Bing Yu et~al.
\newblock The deep ritz method: a deep learning-based numerical algorithm for
  solving variational problems.
\newblock {\em Communications in Mathematics and Statistics}, 6(1):1--12, 2018.

\bibitem{petsc-sf}
Junchao Zhang, Jed Brown, Satish Balay, Jacob Faibussowitsch, Matthew Knepley,
  Oana Marin, Richard~Tran Mills, Todd Munson, Barry~F Smith, and Stefano
  Zampini.
\newblock The {PetscSF} scalable communication layer.
\newblock {\em IEEE Transactions on Parallel and Distributed Systems},
  33(4):842--853, 2021.

\bibitem{cryodrgn}
Ellen~D Zhong, Tristan Bepler, Bonnie Berger, and Joseph~H Davis.
\newblock {CryoDRGN}: reconstruction of heterogeneous cryo-{EM} structures
  using neural networks.
\newblock {\em Nature methods}, 18(2):176--185, 2021.

\end{thebibliography}

\end{document}